\documentclass[runningheads]{llncs}
\usepackage{graphicx}
\usepackage{bbm}
\usepackage[hidelinks]{hyperref}
\usepackage{xcolor}
\usepackage{amsmath}
\usepackage{amssymb}
\usepackage{multirow}
\usepackage{colortbl}
\newcommand{\savefootnote}[2]{\footnote{\label{#1}#2}}
\newcommand{\repeatfootnote}[1]{\textsuperscript{\ref{#1}}}
\begin{document}
\title{Evaluating Adversarial Robustness on Document Image Classification}
%
%
\author{Timothée Fronteau\orcidID{0000-0003-2207-8864} \and
Arnaud Paran\inst{*}\orcidID{0000-0001-9657-2828} \and
Aymen Shabou\orcidID{0000-0001-8933-7053}}
\authorrunning{T. Fronteau et al.}
\institute{DataLab Groupe, Credit Agricole S.A, Montrouge , France \\
\email{fronteau.timothee@gmail.com} \\
\email{\{arnaud.paran, aymen.shabou\}@credit-agricole-sa.fr}\\
\url{https://datalab-groupe.github.io/}\\
\inst{*} Corresponding Author}
\maketitle              
\begin{abstract}
Adversarial attacks and defenses have gained increasing interest 
on computer vision systems in recent years, but as of today, most
investigations are limited to natural images. However, many artificial 
intelligence models actually handle documentary data, which is
very different from real world images. Hence, in this work, we
try to apply the adversarial attack philosophy on documentary data
and to protect models against such attacks. Our methodology is to implement untargeted gradient-based, transfer-based and 
score-based attacks and evaluate the impact of defenses such as
adversarial training, 
JPEG input compression and grey-scale input transformation on the 
robustness of ResNet50 and EfficientNetB0 model architectures. To 
the best of our knowledge, no such work has been conducted by the 
community in order to study the impact of these attacks on the
document image classification task.

\keywords{Adversarial attacks \and Computer vision \and Deep learning.}
\end{abstract}
%
%
%
\section{Introduction}

The democratization of artificial intelligence (AI) and deep 
learning systems has raised public concern about the 
reliability of these new technologies, particularly in terms of 
safety and robustness. These considerations among others have 
motivated the elaboration of a new European regulation, known as 
the \textit{AI Act}, which aims to regulate the use of AI systems.

In a context of substantial increase of incoming customer 
documents in big companies, document classification based on 
computer vision techniques has been found to be an effective 
approach for automatically classifying documents such as ID 
cards, invoices, tax notices, etc. Unfortunately, these 
techniques have demonstrated to be vulnerable to adversarial 
examples, which are maliciously modified inputs that lead to 
adversary-advantageous misclassification by a targeted model  
\cite{szegedy2013intriguing}.

However, although robustness of image classification models 
against several adversarial example generation methods has been 
benchmarked on datasets like ImageNet and Cifar10 
\cite{dong2020benchmarking}, the conclusions of these studies 
might not apply to document image classification. In fact, documents 
have different semantic information than most images: they have 
text, a layout template, often 
a light background and logos. This is why the adversarial
robustness of visual models for document
image classification should be evaluated in an appropriate 
attack setting, and on an appropriate dataset.

In this paper, we evaluate the robustness of two state-of-the-art 
visual models for document classification on the RVL-CDIP dataset 
\cite{harley2015icdar}. Our contributions are as follows:
\begin{itemize}
    \item We establish a threat model that is consistent with the 
    document classification task we study following the taxonomy 
    of \cite{carlini2019evaluating}, and propose a new 
    constraint for generating adversarial examples, adapted to documents.
    \item We evaluate an EfficientNetB0 architecture 
    \cite{tan2019efficientnet} and a ResNet50 architecture 
    \cite{he2016deep}, each one trained two times using 
    different methods, against several adversarial attacks 
    that are representative of the different 
    threat scenarios we can imagine.
\end{itemize}

After giving an overview of related works on adversarial attacks and 
document image classification, we present our experimental data and method. 
Then we describe our experiments and results. Finally, we discuss our 
findings and conclude the work.

\section{Related work}
\subsection{Adversarial attacks on Image classification}

Szegedy et al. \cite{szegedy2013intriguing} are the first ones to 
show that well performing deep neural networks can be vulnerable to 
adversarial attacks, by generating adversarial examples for an 
image classification task. The topic rapidly gained in popularity 
among the deep learning community, which led to the publishing of 
several surveys on the matter.

Following that work, multiple gradient based attacks emerged. Those attacks used projected gradient descent in order to maximize the loss of the model\cite{goodfellow2014explaining}\cite{kurakin2018adversarial}\cite{dong2018boosting}. The idea behind that approach is that when the loss is at a high value, the model will misclassify the data. Carlini and Wagner\cite{carlini2017towards} presented an attack which works in a targeted scenario meaning that we choose the category we want the data to be labeled as. More recently, Sriraman et al\cite{sriramanan2020guided} used graduated optimization and replaced the common crossentropy from the loss by a margin between the probability of the correct label and the probability of the adversarial label. That replacement leads to better results than using crossentropy.

In a black box setting, successfully training a surrogate model that will be attacked proved being a decent approach\cite{zhang2022towards}

With the goal of hiding adversarial attacks in the details of pictures, Jia et al\cite{jia2022exploring} didn't attack in pixel space but rather in discrete cosine transform space on KxK square patches of the image. The idea behind that approach is that the attack will merge better with the original image.

Liu et al\cite{liu2022practical} perfected an approach which tries to find optimal perturbations to initialize adversarial search by using adaptative methods. That initialization allows us to explore better the space of adversarial candidates.

Ensemble methods try to attack multiple models at once. One issue is that the models can be very different. To tackle that issue, SVRE\cite{xiong2022stochastic} uses a variance reduction strategy which allows gradient descent to perform better.

The field of universal perturbations explored attacks which can fool most of the images of the dataset the attack is performed on\cite{moosavi2017universal}.

Chakraborty et al. \cite{chakraborty2021survey} and Machado et al. \cite{machado2021adversarial}
categorize adversarial attacks into threat frameworks depending on 
the attack surface, capabilities and on the adversary’s goal, and 
present representative methods for each attack threat. The taxonomy 
of the first article covers adversarial attacks in the vast domain 
of computer vision, while the second one focuses on image 
classification.

This threat framework, or threat model, is important to compare 
attacks. In particular, the authors distinguish white-box to 
black-box attacks. In the first kind of attack, the adversary has 
complete knowledge of the target model and its parameters. In the 
second kind, the attacker has partial or no knowledge of the model 
and of its training data. Depending on the attacker’s knowledge, 
different attack approaches can be excluded by the threat model, 
which offers the possibility to compare attacks within a threat 
model.

Both articles also provide a state of the art of the main defense 
strategies. However, among the many possibilities that explain the 
existence of adversarial examples, the scientific community 
commonly accepts none and it is challenging to design effective 
defense strategies \cite{machado2021adversarial}. In order to help 
in this process, Carlini et al.\cite{carlini2019evaluating} gave a thorough 
set of recommendations on how to evaluate attack methods and 
defense strategies depending on the attack threat. 

Dong et al. \cite{dong2020benchmarking} selected several adversarial example 
generation methods that cover the spectrum of categories listed 
by the previous articles, and benchmarked the robustness of defense 
models against these attacks on a classification task. They used 
the CIFAR-10 \cite{krizhevsky2009learning} and ImageNet 
\cite{deng2009imagenet} datasets of labelled pictures. They 
implemented 19 attack methods and 8 defense models using the 
PyTorch framework and published a library of example generation 
methods in Python and robustness evaluation pipelines on GitHub \savefootnote{ares}{Ares robustness library URL:\url{https://github.com/thu-ml/ares}}. 
Another library, CleverHans, implements five of these attacks 
using the Tensorflow framework \cite{papernot2018cleverhans}.

\subsection{Document image classification}
Few open datasets are accessible to evaluate and compare methods 
for document image classification. The RVL-CDIP dataset 
\cite{harley2015evaluation} is at the time a commonly used dataset
to compare deep learning methods for classifying 
documents\savefootnote{paperswithcodedataset}{Meta AI. 
\textit{Document Image Classification}. June 2020. URL: 
\url{https://paperswithcode.com/task/document-image-classification}}. 
The most recent approaches that obtain 
state-of-the-art accuracy are Transformer-based approaches that 
make use of visual, textual and layout information, like DocFormer
\cite{appalaraju2021docformer} or LayoutLMv2 
\cite{xu2020layoutlmv2}. These methods make use of hundreds of 
thousands of parameters to classify one document, and require the use of an OCR 
as a preprocessing step to extract the text from documents.
For these two reasons, their inference time is longer than deep learning 
models based on convolutional layers. Also, those models are very heavy compared to convolutional models that only use the visual 
information of documents while reaching good accuracy, although a 
bit lower \cite{ferrando2020improving}. With the notable 
exception of VGG16 \cite{simonyan2014very}, these models have an 
order of magnitude smaller number of parameters 
\cite{ferrando2020improving,afzal2017cutting}.
For those reasons, lightweight convolutional models will be favored unless there is a real need of the extra power of multimodal models.
That's why we decided to focus on visual convolutional models as this case
will cover more real life applications in several industries.

These models have been trained and optimized to obtain 
state-of-the-art accuracy, without consideration for their 
adversarial robustness. However, accurate models are not 
necessarily robust to adversarial examples, indeed they often 
perform poorly in comparison to humans. Therefore, evaluating 
adversarial robustness is a tool for understanding where these 
systems fail and make progress in deep learning 
\cite{carlini2019evaluating}.

\section{Method and Data}
In our attempt to evaluate the adversarial robustness of common 
document image classification methods, we carefully selected a 
proper dataset for document classification, a range of threat 
models in which we want to perform our evaluation, and a few attack 
methods and defense models that suit these settings.

\subsection{Data}
The \textit{RVL-CDIP} dataset \cite{harley2015evaluation} is an 
open dataset containing 400,000 black-and-white document images 
split into 16 categories among which we can find "news article", 
"advertisement", "email" or "handwritten", for example. We 
subsampled the images into images of shape 240x240 pixels, and 
duplicate the pixel values into the three RGB channels so that 
they fit our models, which are presented below. The images are 
therefore of shape 240x240x3.

The dataset is already split into a training set, a validation set 
and a test set of 360,000, 40,000 and 40,000 documents 
respectively, saved in TIFF image format. We randomly selected 1000
documents from the test set to perform our evaluations, containing 54 
to 68 examples for each of the 16 classes.

\subsection{Threat Model}
In order to have correct metrics, and to assess the validity of our 
approach in a real-world context, it is essential that we define a 
threat model using a precise taxonomy. We define this threat model 
according to Carlini et al \cite{carlini2019evaluating} by defining the goals,
capabilities, and knowledge of the target AI system, that the 
adversary has.

Let $C(x) : X \to Y$ be a classifier, and $x \in X \subset 
\mathbb{R}^d$ an input to the classifier. Let $y^{true} \in Y 
= \{1, 2, ..., N\}$ be the ground truth of the input $x$, i.e. 
the true class label of $x$ among $N$ classes. We call $y^{pred} 
= C(x)$ the predicted label for $x$. 

In our study, we perform \textit{untargeted attacks}, ie. the 
adversary's goal is to generate an adversarial example $ x^{adv} $ 
for $x$ that is misclassified by $C$. This goal is easier to reach
than the one of a targeted attack, where $ x^{adv} $ should be 
classified as a predefined $y^{adv} \neq y^{true}$ to be considered 
adversarial. $ x^{adv}$ should also be as 
\textit{optimal} as possible, which means that it fools the model 
with high confidence in the wrong predicted label, with an input 
$x^{adv}$ that is indistinguishable from $x$ by a human eye
\cite{machado2021adversarial}. 

Formally, we can define the \textit{capability} of the adversary by 
defining a \textit{perturbation budget} $\varepsilon$ so that $\Vert 
x - x^{adv} \Vert_p < \varepsilon$ where $\Vert \cdot \Vert_p$ is the 
$L_p$ norm with $p \in \{2, \infty\}$. For some attack methods, the 
capability of the adversary can also be described with the 
\textit{strength} of the attack, that we define as the number of 
queries an adversary can make to the model. 

\subsection{Attack methods}
Here we present the nine attack methods we used for generating 
adversarial examples. We selected them in order to cover a large 
spectrum of the capability settings an adversary can have. We cover 
white-box to black-box attacks, namely gradient-based, transfer-based
and score-based attacks, under the $L_{\infty}$ and the $L_2$ norms.

\begin{figure}
\begin{center}
\includegraphics[height=0.8\textheight]{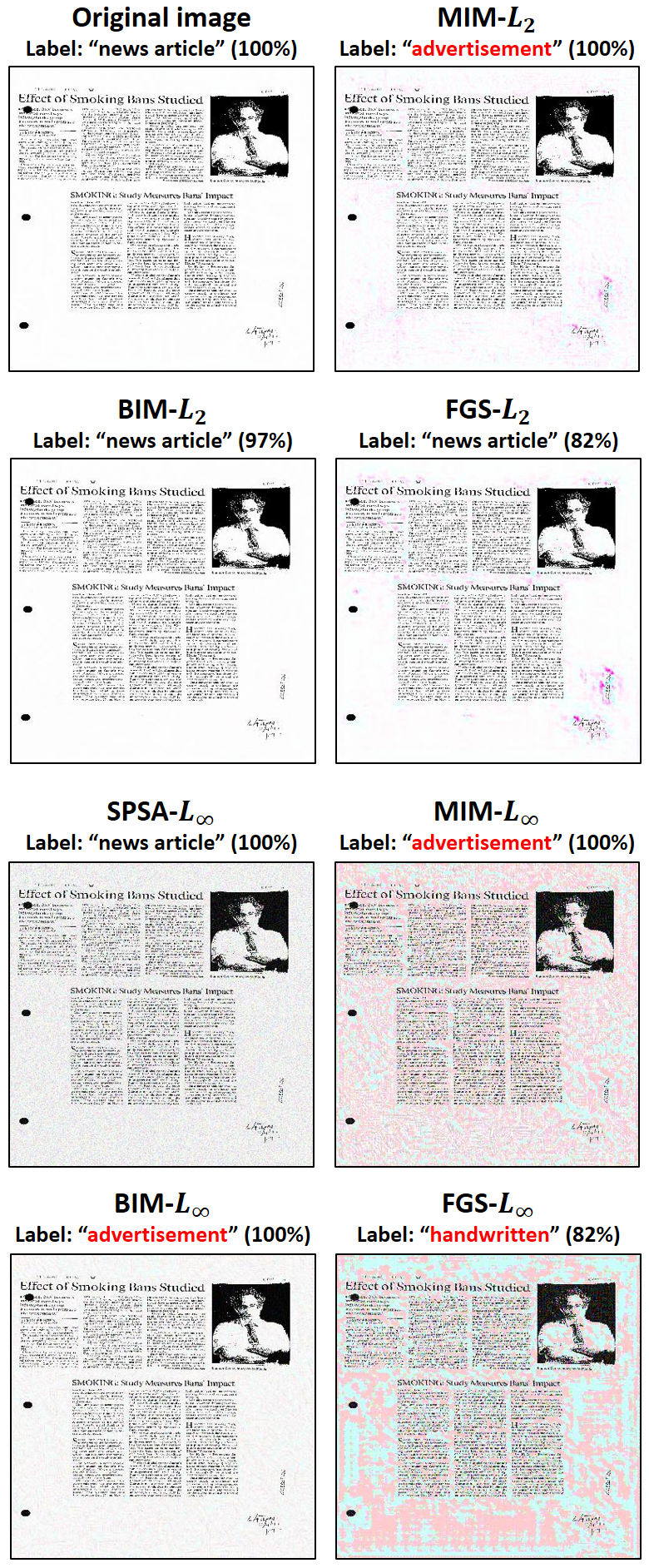}
\end{center}
\caption{\textbf{An original image of the RVL-CDIP dataset along with 
perturbed images generated with gradient-based and score-based attacks 
against EfficientNet with JPEG defense.} The true label of the image is 
"news article". We set the perturbation budget to 0.2 for the attacks 
under $L_\infty$ norm, and to 20 for the others. On top of each image 
figures the label predicted by the model, along with the 
confidence of the model in its prediction. The images with red labels
are successful attack images.} \label{fig1}
\end{figure}

Examples of the adversarial images computed with gradient-based and
score-based attacks are presented in Figure \ref{fig1}.

\subsubsection{Gradient-based attacks.}
Considering that the adversary has access to all parameters and weights 
of a target model, we can generate a perturbation for $x$ by using the 
gradient of a loss function computed with the model parameters. The Fast
Gradient Method (FGM) \cite{goodfellow2014explaining}, the Basic Iterative 
Method (BIM) \cite{kurakin2018adversarial} and the Momentum Iterative Method 
(MIM) \cite{dong2018boosting} are three gradient-based methods we evaluated
in this paper. BIM and MIM are recursive versions of FGM, which requires
only one computation step.

The three attacks have a version well suited for generating perturbations 
under the $L_{\infty}$ constraint and another that suits the $L_2$ constraint,
so we evaluated both versions for a total of six different attacks.
When necessary, we differentiate the $L_{\infty}$ and the $L_2$ versions of
the attacks with the suffix -$L_{\infty}$ and -$L_2$ respectively.

\subsubsection{Transfer-based attacks.}
Adversarial images generated with gradient-based attacks on a substitute 
model have a chance of also leading the target model to misclassification.
To evaluate the extent of this phenomenon of transferability 
\cite{papernot2016practical}, we used the perturbed examples generated with 
robust models to attack a target model in a transfer-based setting. The
examples are generated under $L_{\infty}$ constraint with FGM and MIM, 
which is a variation of BIM designed to improve the transferability of 
adversarial examples. Note that the adversary has access to the training 
set of the target model to train the substitute model.

\subsubsection{Score-based attack.}
Considering that the adversary has only access to the prediction array
of a target model and can query it a maximum number of times defined as 
the attack strengh, we can design the so-called score-based attacks. The
Simultaneous Perturbation Stochastic Approximation method 
(SPSA) \cite{uesato2018adversarial} is one of them. We evaluate it 
under $L_{\infty}$ constraint.

\subsection{Models and Defenses}
\subsubsection{Model backbones.}

We conducted our experiments on two visual model architectures that 
currently perform best on RVL-CDIP with less than a hundred million of
parameters, as far as we know \savefootnote{paperswithcode}{https://paperswithcode.com/sota/document-image-classification-on-rvl-cdip}: an 
EfficientNetB0 and a ResNet50. According to the method of 
Ferrando et al.\cite{ferrando2020improving}, who observed the best accuracy of 
both models on our classification task, we initialized these models 
with weights that have been pre-trained on the ImageNet 
classification task. 

\subsubsection{Reactive defenses.}
We implemented preprocessing steps, that precede the model 
backbone, with the aim of improving their robustness. The first of these 
reactive defenses we implemented is the JPEG preprocessing. We compress the document image into the JPEG format, then decompress each image that is given as input 
of the model backbone (\textbf{JPEG} defense). This transformation has 
been identified by Dsiugaite et al. \cite{dziugaite2016study} as a factor of robustness 
against adversarial attacks. This transformation is all the more 
interesting because it is widely used for industrial applications.

The grey-scale transformation is an other low-cost preprocessing step 
that we evaluated. The model backbones require that we use three color 
channels, but document images may contain less semantic information 
accessible via the colors of the image. As a matter of fact, the 
RVL-CDIP dataset is even composed of grey-scale images. Therefore, 
constraining the input images before the model backbone by averaging 
the values of each RGB pixel (\textbf{Grey} defense) does not affect 
the test accuracy of the models, but might improve adversarial robustness.

\subsubsection{Proactive defense.}
Some training methods have proven to improve robustness against 
adversarial attacks on image classification tasks 
\cite{dong2020benchmarking}. In this article, we compare the natural 
fine-tuning proposed in Ferrando et al.\cite{ferrando2020improving} with the 
adversarial training method suggested in Kurakin et al.SS
\cite{kurakin2016adversarial}, which has the advantage of being 
very easy to implement. 

In this last method, the fine-tuning step is performed using the 
adversarial examples generated with BIM (\textbf{Adversarial} 
defense). One adversarial example is 
generated for each input of the training set and for each epoch of 
the training. We use the same learning rate scheduling as for the 
natural training method. This means that the attacks are performed against the training model and not only against the naturally trained model.


\section{Experiments and Results}
Similar to Dong et al.\cite{dong2020benchmarking}, we selected two distinct 
measures to assess the robustness of the models under 
attack, defined as follows. Given an attack method $A$ that generates 
an adversarial example $x^{adv} = A(x)$ for an input $x$, the 
accuracy of a classifier $C$ is defined as
$ Acc(C, A) = \frac{1}{N}\sum_{i=1}^N \mathbbm{1} (C(A(x_i)) = 
y_i)\mbox{,} $
where $\{x_i, y_i\}_{1 \leq i \leq N}$ is the test set and $\mathbbm{1}(\cdot)$ 
is the indicator function. The attack success rate of an untargeted 
attack on the classifier is defined as
$ Asr(C, A) = \frac{1}{M}\sum_{i=1}^N \mathbbm{1}(C(x_i) = y_i	\land 
C(A(x_i)) \ne y_i)\mbox{,} $
where $M = \Sigma_{i=1}^N\mathbbm{1}(C(x_i) = y_i)$. 

On Figures \ref{fig2}, \ref{fig3} and \ref{fig4}, we draw curves of 
accuracy vs. perturbation budget of each model under the nine attacks
we evaluated. On the other hand, the test accuracy of each model 
under no attack is rendered on Table \ref{tab1}, and ranges from 
86.1\% to 90.8\%.

\begin{table}
\begin{center}
\begin{tabular}{ |c||c|c|  }
\cline{2-3}
\multicolumn{1}{c||}{} & \multicolumn{2}{ c |}{\textbf{Model Backbone}} \\
 \hline
\textbf{Defense Method} & EfficientNet & ResNet\\
 \hline
 \hline
No Defense                  &  \textbf{90.8} & \textbf{89.0} \\
Grey                        &  \textbf{90.8} & \textbf{89.0} \\
JPEG                        &  86.8          & 86.1          \\
Adversarial                 &  89.0          & 87.3          \\
Grey + JPEG + Adversarial   &  89.3          & 87.2          \\
 \hline
\end{tabular}
\caption{\textbf{Test accuracy of each model.} The test accuracy has been computed on the same 1000 examples as for all experiments.} \label{tab1}
\end{center}
\end{table}

\subsection{Adversarial Robustness under Gradient-based Attack}
\begin{figure}
\includegraphics[width=\textwidth,height=4cm]{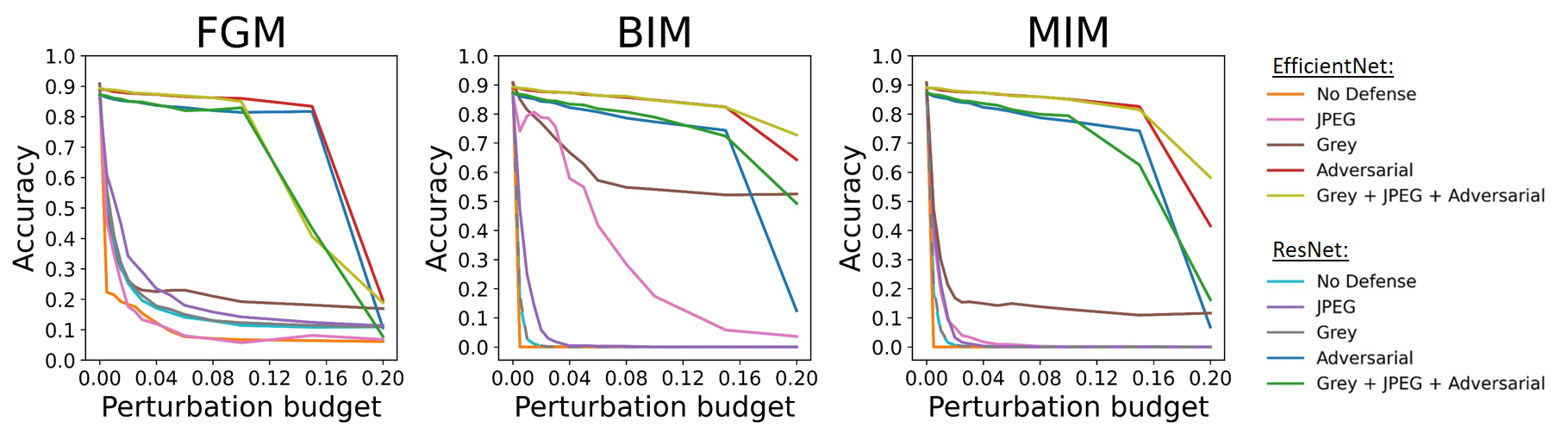}
\caption{\textbf{The accuracy vs. perturbation budget curves of 
gradient-based attacks under $L_\infty$ norm.}} \label{fig2}
\end{figure}

\begin{figure}
\includegraphics[width=\textwidth,height=4cm]{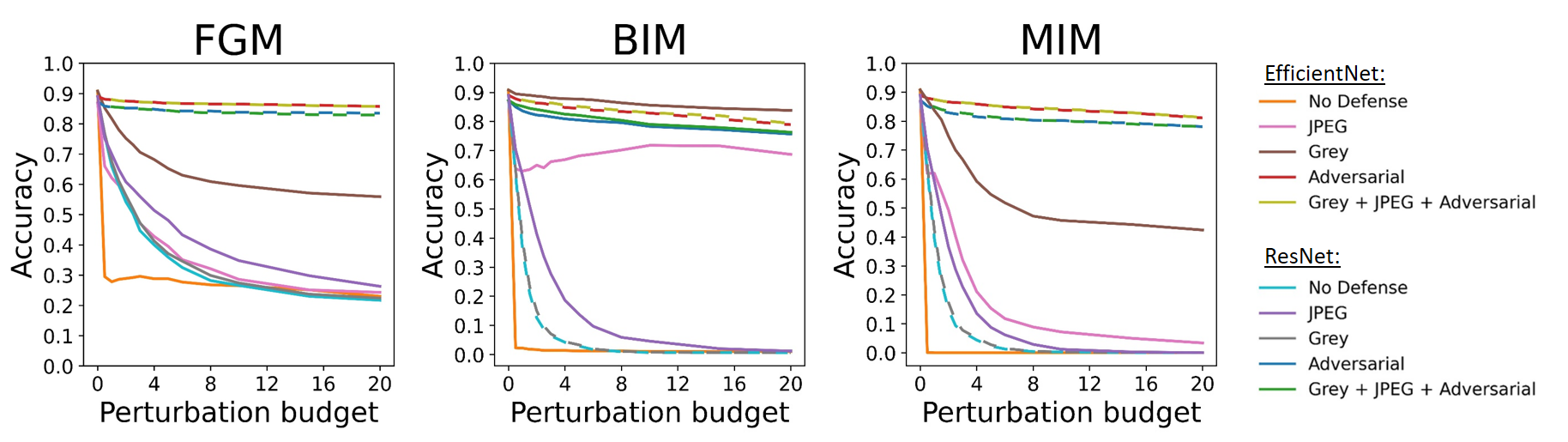}
\caption{\textbf{The accuracy vs. perturbation budget curves of 
gradient-based attacks under $L_2$ norm.}} \label{fig3}
\end{figure}

We performed the gradient-based attacks under $L_{\infty}$ and 
$L_{2}$ constraints
for perturbation budgets within the ranges of 0.005 to 0.20, 
and of 0.5 to 20 respectively. We display the results 
in Figure \ref{fig2} and Figure \ref{fig3} respectively,
and render them in Table \ref{tab2}. 
For BIM and MIM, we set the attack strength to 20 queries to 
the target model.

\begin{table}
\begin{center}
\begin{tabular}{ |c||c||c|c|c||c|c|c||c|  }
\cline{3-9}
 \multicolumn{2}{c||}{} & \multicolumn{7}{ c |}{Attack Method} \\
 \hline
 \multirow{2}*{Backbone} & \multirow{2}*{Defense Method} & FGM & BIM & MIM & FGM & BIM & MIM & SPSA \\
 \cline{3-8}
 & & \multicolumn{3}{|c||}{X-$L_\infty$} &\multicolumn{3}{|c||}{X-$L_2$ } & -$L_\infty$ \\
 \hline
 \hline
  \multirow{5}*{\rotatebox[origin=c]{60}{EfficientNet}} &
       No defense                & 18.3  & 00.0  & 00.0 & 28.9 & 01.7 & 00.0 & 02.5 \\
& Grey                      & 26.2  & 76.9  & 16.8 & 75.2 & \textbf{88.7} & 74.8 &  -   \\
 & JPEG                      & 17.6  & 78.8  & 06.7 & 54.4 & 65.0 & 49.3 & 79.1 \\
 & Adversarial               & 87.7  & 87.7  & 87.7 & \textbf{87.5} & 86.3 & \textbf{86.6} & \textbf{88.0} \\
 & Grey + JPEG + Adversarial & \textbf{88.1}  & \textbf{87.9}  & \textbf{87.9} & 87.4 & 86.6 & \textbf{86.6} &  -   \\
 \hline
 \multirow{5}*{\rotatebox[origin=c]{60}{ResNet}}
 &     No defense                & 25.3  & 00.3   & 00.6  & 54.2 & 12.6 & 14.9 & 39.7 \\
 & Grey                      & 26.5  & 00.4   & 00.7  & 56.4 & 15.0 & 17.4 &  -   \\
 & JPEG                      & 34.2  & 05.8   & 03.2  & 60.8 & 41.4 & 36.6 & 47.9 \\
 & Adversarial               & 85.0  & 84.3   & 84.4  & 85.2 & 82.2 & 82.9 & \textbf{85.1} \\
 & Grey + JPEG + Adversarial & \textbf{85.1}  & \textbf{85.1}   & \textbf{85.1}  & \textbf{85.3} & \textbf{84.1} & \textbf{83.5} &  -   \\
 \hline
\end{tabular}
\caption{\textbf{The accuracy models under several gradient-based and score-based attacks.} 
The attacks have been computed with 
a perturbation budget of 0.02 for attacks under $L_\infty$ constraint, 
and of 2 for attacks under $L_2$ constraint.} \label{tab2}
\end{center}
\end{table}

First we can see that the accuracy of undefended models drops 
drastically from more than 88\% to less than 0.6\% under 
BIM-$L_{\infty}$ and MIM-$L_{\infty}$ attacks with $\varepsilon=0,02$, 
and to less than 14.9\% under BIM-$L_{2}$ and MIM-$L_{2}$ attacks
with $\varepsilon=2$. We see that the JPEG and Grey defenses don't
necessarily improve the adversarial robustness of target models,
and when they do, the improvements are inconsistent depending on 
the attack and the considered model backbone. For example, 
with $\varepsilon=0.02$ against BIM-$L_{\infty}$,
the accuracy of EfficientNet with Grey defense goes up to 76.9\% 
from 0\%, while the accuracy of ResNet only improves by 0.1 
point from 0.3\%, and the accuracy of EfficientNet under 
MIM-$L_{\infty}$ goes up to 16.8\% from 0\%.

The adversarial training, on the other hand, improves consistently for 
all model backbones and under all evaluated attacks the robustness 
of targeted models. In fact, accuracies of adversarially 
trained models only drop of 1.9 points in average against attacks
under $L_{\infty}$ constraint with $\varepsilon=0.02$, and
of 2.9 points in average against attacks
under $L_{2}$ constraint with $\varepsilon=2$.

\subsection{Adversarial Robustness under Transfer-based Attack}
\begin{figure}
\includegraphics[width=\textwidth]{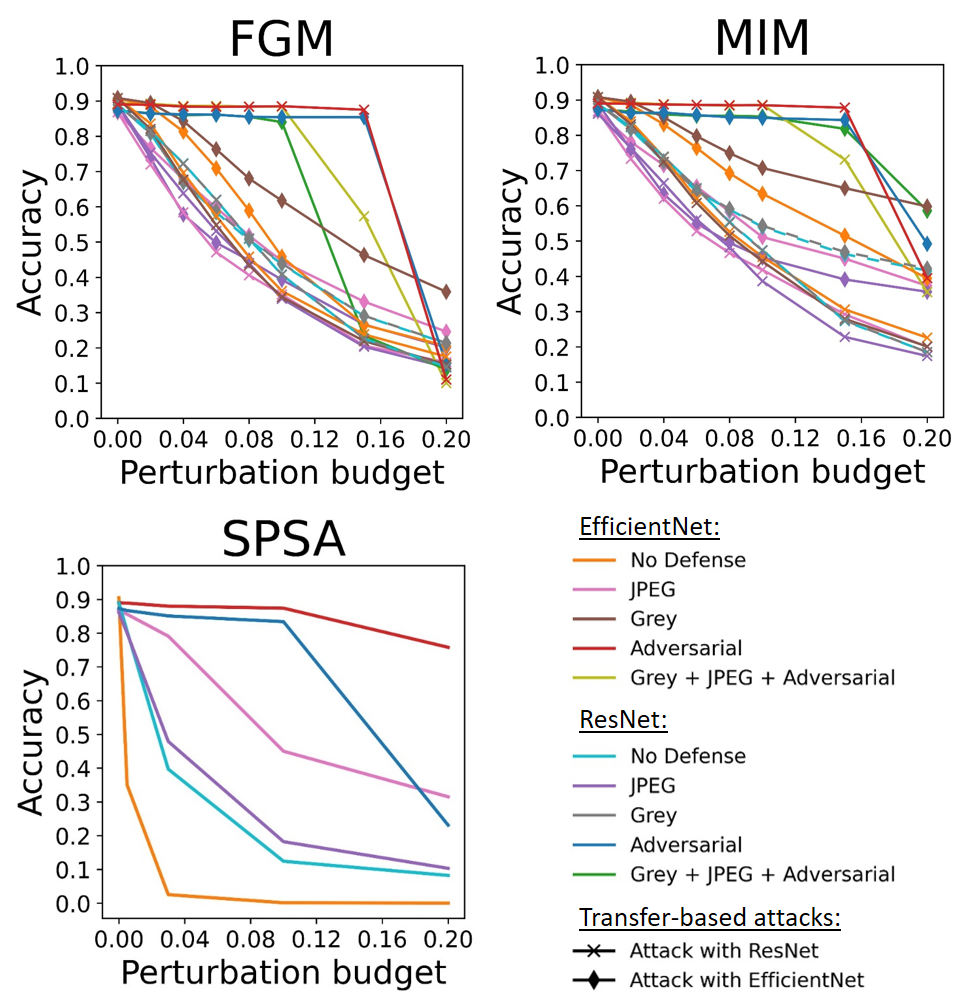}
\caption{\textbf{The accuracy vs. perturbation budget curves of 
transfer-based attacks (FGM, MIM) and a score-based attack (SPSA) under $L_\infty$ norm.}} \label{fig4}
\end{figure}

We generated perturbed examples with FGM-$L_{\infty}$ and 
MIM-$L_{\infty}$ by using the adversarially 
trained model backbones EfficientNet and ResNet as subtitute models to attack 
every other model in a transfer-based setting, and render the results in 
Figure \ref{fig4} and Table \ref{tab3}. 

\begin{table}
\begin{center}
\begin{tabular}{ |c||c||c|c||c|c|  }
 \hline
 \multirow{4}*{Backbone} & \multirow{4}*{Defense Method} & \multicolumn{4}{|c|}{\textbf{Substitute Model}} \\
 \cline{3-6}
 & & \multicolumn{2}{|c||}{EfficientNet} & \multicolumn{2}{|c|}{ResNet} \\
 \cline{3-6}
 \cline{3-6}
 & & \multicolumn{4}{|c|}{\textbf{Method}} \\
\cline{3-6}
& & FGM-$L_\infty$ & MIM-$L_\infty$ & FGM-$L_\infty$ & MIM-$L_\infty$ \\
 \hline
 \hline
  \multirow{5}*{\rotatebox[origin=c]{60}{EfficientNet}} &
       No defense                & \cellcolor[gray]{0.8}  & \cellcolor[gray]{0.8}  & 83.6  & 83.9  \\
 & Grey                      & \cellcolor[gray]{0.8}  & \cellcolor[gray]{0.8}  & 82.0  & 83.1  \\
 & JPEG                      & \cellcolor[gray]{0.8}  & \cellcolor[gray]{0.8}  & 72.0  & 73.3  \\
 & Adversarial               &   \cellcolor[gray]{0.8}   &   \cellcolor[gray]{0.8}   & 88.9  & 88.9  \\
 & Grey + JPEG + Adversarial &   \cellcolor[gray]{0.8}   &   \cellcolor[gray]{0.8}   & \textbf{89.2}  & \textbf{89.2}  \\
 \hline
 \multirow{5}*{\rotatebox[origin=c]{60}{ResNet}}
 &     No defense                & 80.8  & 81.7  & \cellcolor[gray]{0.8}  & \cellcolor[gray]{0.8}  \\
 & Grey                      & 80.9  & 81.8  & \cellcolor[gray]{0.8}  & \cellcolor[gray]{0.8}  \\
 & JPEG                      & 73.8  & 78.1  & \cellcolor[gray]{0.8}  & \cellcolor[gray]{0.8}  \\
 & Adversarial               & 86.4  & 86.3  &   \cellcolor[gray]{0.8}   &   \cellcolor[gray]{0.8}   \\
 & Grey + JPEG + Adversarial & \textbf{86.5}  & \textbf{86.7}  &   \cellcolor[gray]{0.8}   &   \cellcolor[gray]{0.8}   \\
 \hline
\end{tabular}
\caption{\textbf{The accuracy models under several transfer-based attacks 
under $L_\infty$ constraint, with a perturbation budget of 0.02.}
Target models have been attacked with two substitute models: an adversarially trained 
EfficientNet and an adversarially trained ResNet. As we consider transfer attacks we didn't rerun attacks towards the same model.} \label{tab3}
\end{center}
\end{table}

We can observe that undefended models are vulnerable to adversarial images 
generated with the more robust substitute models (up to 36 points of accuracy 
decrease), while adversarially trained models stay robust (less than 4 
points of accuracy decrease) for a perturbation budget $\varepsilon=0.10$.
Interestingly, the JPEG defense seems to increase the vulnerability of models 
to adversarial attacks: the curves of EfficientNet and ResNet with JPEG defense 
stay under the curves of undefended EfficientNet and ResNet when they are 
attacked by the same perturbed examples. The Grey defense doesn't change 
the accuracy of the naturally trained ResNet by more than 2 points, while 
the Grey defense on
the naturally trained EfficientNet has inconsistent behaviour depending on 
the adversary's model, improving robustness against the EfficientNet 
substitute model but reducing it against ResNet model. 

\subsection{Adversarial Robustness under Score-based Attack}
The SPSA attack requires to set several parameters. According to the taxonomy of Uesato et al.
\cite{uesato2018adversarial}, we perform 20 iterations of the method as for BIM and 
MIM, and for each iteration, we compute a batch of 10 images with 40 pixel perturbations 
each. Therefore, it was 40 times longer to compute an adversarial example with SPSA than 
with BIM or MIM, as we could observe. With such parameters, we can reduce the 
robustness of every model in a 
similar way than with gradient-based attacks (see Figure \ref{fig4} and Table \ref{tab3}). 
In fact, in average, models with no defense have 2.5\% and 39.7\% of accuracy under SPSA 
attack with a perturbation budget $\varepsilon=0.10$, with 
EfficientNet and ResNet backbones respectively.
With JPEG defense, they still have 79.1\% and 47.9\% of accuracy respectively, and 
the accuracy goes up to 88.0\% and 85.1\% for adversarially trained models.
It shows that naturally trained document 
classification models can be vulnerable to score-based attacks if the threat model 
(the attack strength in particular) allows it.

\section{Conclusion and Future Works}

As expected, a convolutional model such as an EfficientNetB0 or a ResNet50 trained without 
any strategy to improve its robustness, is very sensitive to optimal adversary examples 
generated with gradient-based white-box untargeted attacks, which constitute attacks performed 
in the best threat model for an adversary.
Restraining the values of pixels 
so that the images are grey-scale doesn't improve much the 
robustness of the models. Compressing and then decompressing input 
images of a model using JPEG protocol improves the model 
robustness against adversarial images, but not consistently.  

On the other hand, the 
adversarial training of both models using the method of Kurakin et al.
\cite{kurakin2016adversarial}, strongly improves robustness of both models. This 
training method is quite easy to implement and does not affect a lot the test accuracy of 
models on our classification task. Therefore, this method seems far more effective on a 
document classification task than against an image classification task, as we can see in Dong et al.
\cite{dong2020benchmarking}.

The black-box attack we evaluated generates blurry examples that appear darker than 
legitimate document images. It seems to be the case with other black-box attacks, since the 
perturbation is generated randomly using probabilistic laws that don't take into account 
the fact that document images are brighter than other images \cite{dong2019efficient}. 

There are many ways to improve the robustness of a model that would only use the visual 
modality of a document \cite{chakraborty2021survey,machado2021adversarial}. However, 
state-of-the-art approaches to document classification take advantage of other information 
modalities, such as the layout of the document, and the text it contains 
\repeatfootnote{paperswithcode}. Therefore, after this work on evaluating the robustness of 
visual models, it would be interesting to evaluate the transferability of the generated 
examples to a multimodal model such as DocFormer or LayoutLMv2, which use optical character 
recognition (OCR) and transformer layers. Furthermore, we could explore the 
possibility of designing adversarial attacks to which these models are more sensitive, for 
example by targeting OCR prediction errors \cite{song2018fooling} that affect the textual 
modality and may also affect the robustness of such models \cite{zhang2020adversarial}. Dealing with the added modality given by text means that more approaches can be explored attacking only one modality or both.

Other defense mechanisms could be explored too, for example training a model to detect adversarial examples is a common approach. We did not favor that one as it implies using multiple models which makes inference longer but future works could explore that possibility.

On the other hand, attacks specific to document data could be thought of, for example changing the style of the text by changing the font, the size, using bold or italic characters. That would imply a much more complex pipeline to find those adversarial examples but those attacks can benefit from being able to generate perturbations of large amplitude which is a case we did not explore here as our perturbations were bounded in norm.

In conclusion, with this paper, we open up the path to adversarial attacks on documentary data as this issue is a major concern and has not been studied much in the litterature. Our first conclusions are reassuring as adversarial training works on the attacks we explored but further work is needed to explore other attack and defense scenarios.

\bibliographystyle{splncs04}
\bibliography{bibliography}

\end{document}